\documentclass[acmtog]{acmart}
\acmSubmissionID{263}

\newcommand{\ourapproach}{MECo}

\usepackage{multirow}
\usepackage[ruled]{algorithm2e} 

\SetAlFnt{\small}
\SetAlCapFnt{\small}
\SetAlCapNameFnt{\small}
\SetAlCapHSkip{0pt}

\copyrightyear{2025} 
\acmYear{2025} 
\setcopyright{acmlicensed}\acmConference[SIGGRAPH Conference Papers '25]{Special
 Interest Group on Computer Graphics and Interactive Techniques Conference
 Conference Papers }{August 10--14, 2025}{Vancouver, BC, Canada}
 \acmBooktitle{Special Interest Group on Computer Graphics and Interactive
 Techniques Conference Conference Papers (SIGGRAPH Conference Papers '25), August
 10--14, 2025, Vancouver, BC, Canada}
 \acmDOI{10.1145/3721238.3730611}
 \acmISBN{979-8-4007-1540-2/2025/08}

\newcommand{\cbhb}[1]{\textcolor[rgb]{0.00,0.00,0.0}{#1}}

\newcommand{\comments}[1]{}

\begin{document}

\title{Motion-example-controlled Co-speech Gesture Generation Leveraging Large Language Models}

\author{Bohong Chen}
\email{bohongchen@zju.edu.cn}
\orcid{0009-0007-1036-7737}
\affiliation{
\institution{State Key Lab of CAD\&CG, Zhejiang University}
\city{Hangzhou}
\country{China}
}

\author{Yumeng Li}
\email{yumeng.li@zju.edu.cn}
\orcid{0009-0007-6558-4165}
\affiliation{
\institution{State Key Lab of CAD\&CG, Zhejiang University}
\city{Hangzhou}
\country{China}
}

\author{Youyi Zheng}
\email{youyizheng@zju.edu.cn}
\orcid{xxx}
\affiliation{
\institution{State Key Lab of CAD\&CG, Zhejiang University}
\city{Hangzhou}
\country{China}
}

\author{Yao-Xiang Ding}
\email{dingyx.gm@gmail.com}
\orcid{xxx}
\affiliation{
\institution{State Key Lab of CAD\&CG, Zhejiang University}
\city{Hangzhou}
\country{China}
}

\author{Kun Zhou}
\authornote{Corresponding author}
\email{kunzhou@acm.org}
\orcid{0000-0003-4243-6112}
\affiliation{
\institution{State Key Lab of CAD\&CG, Zhejiang University}
\city{Hangzhou}
\country{China}
}

\begin{teaserfigure}
  \centering
  \includegraphics[width=1.0\linewidth]{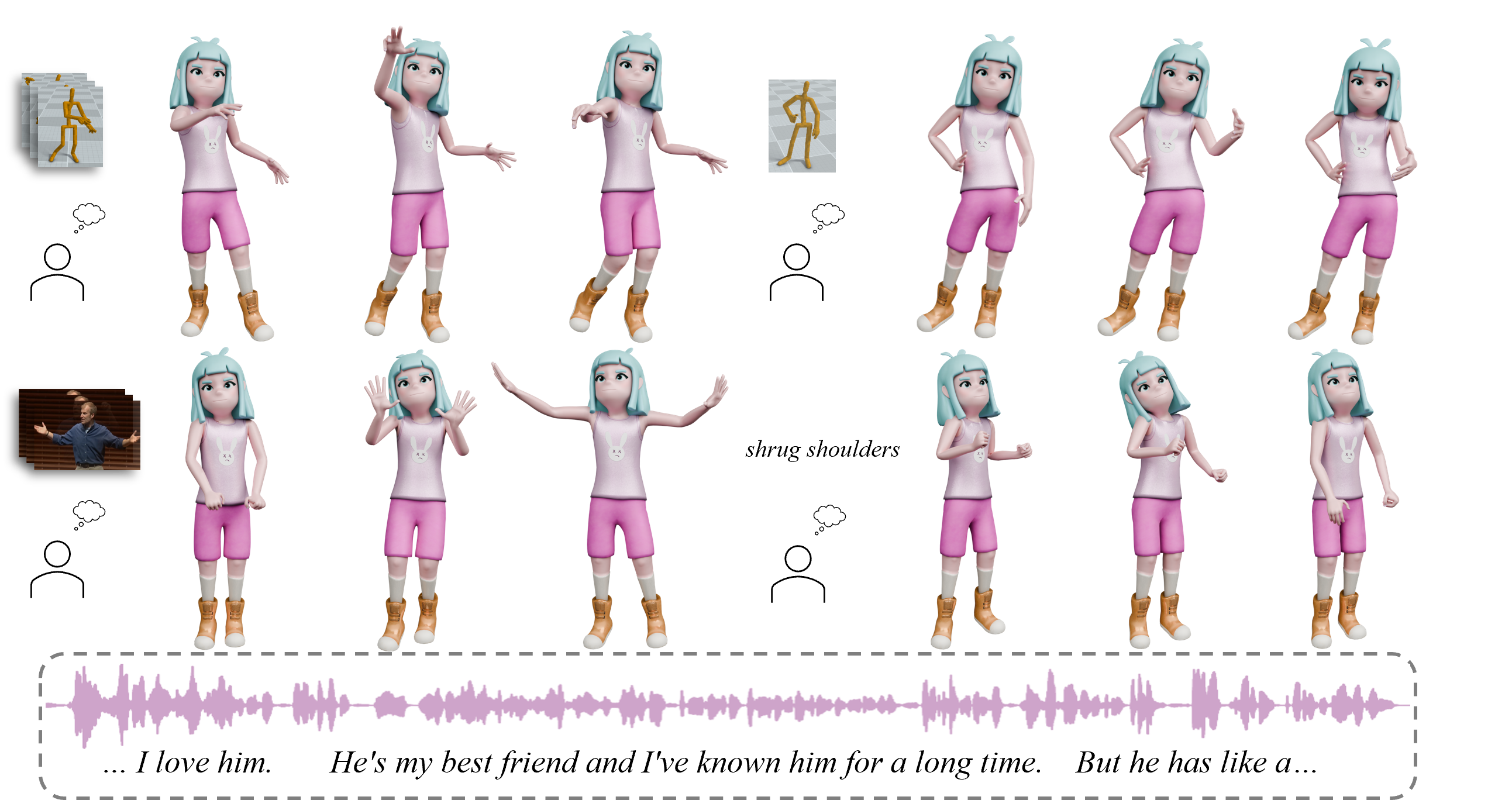}
  \caption{
 Given a motion example and a speech audio clip, our method generates vivid co-speech gestures. Motion examples can be a motion clip, a single pose, a human video, or even a text prompt. The four gestures above are generated by the same speech and four different motion examples.
 The character model is from Adobe Mixamo. 
  }
  \label{fig:teaser}
\end{teaserfigure}

\begin{abstract}
The automatic generation of controllable co-speech gestures has recently gained growing attention. While existing systems typically achieve gesture control through predefined categorical labels or implicit pseudo-labels derived from motion examples, these approaches often compromise the rich details present in the original motion examples. We present \ourapproach, a framework for motion-example-controlled co-speech gesture generation by leveraging large language models (LLMs). Our method capitalizes on LLMs' comprehension capabilities through fine-tuning to simultaneously interpret speech audio and motion examples, enabling the synthesis of gestures that preserve example-specific characteristics while maintaining speech congruence. Departing from conventional pseudo-labeling paradigms, we position motion examples as explicit query contexts within the prompt structure to guide gesture generation. Experimental results demonstrate state-of-the-art performance across three metrics: Fr\'echet Gesture Distance (FGD), motion diversity, and example-gesture similarity. Furthermore, our framework enables granular control of individual body parts and accommodates diverse input modalities including motion clips, static poses, human video sequences, and textual descriptions. Our code, pre-trained models, and videos are available at  https://robinwitch.github.io/MECo-Page.
\end{abstract}

\ccsdesc[500]{Computing methodologies~Motion processing}
\ccsdesc[500]{Computing methodologies~Computer graphics}

\keywords{co-speech motion generation, motion tokens, text-to-motion, multimodal control}

\maketitle

\comments
{
\renewcommand{\thefootnote}{\fnsymbol{footnote}}

\footnotetext[2]{This is the author's version of the work. It is posted here for your personal use. Not for redistribution.}
}

\section{Introduction}
Gestures are the spontaneous and stylized movements of arms, hands and feet that occur while people talk. Just as people have habitual verbal expressions that unconsciously appear in their speech, everyone has their own set of characteristic gestures that they consistently use. These co-speech gestures constitute an essential component of human communication, making the generation of natural and style-appropriate gestures crucial for virtual avatars and digital humans in computer graphics and animation.

Deep learning has become the dominant approach for co-speech gesture generation, yet existing systems often lack fine-grained control mechanisms to effectively translate user intent into precise outputs~\cite{Ao2023GestureDiffuCLIP,ghorbani2022zeroeggs}. Current controllable methods fall into two categories: label-based and example-based. Label-based methods rely on predefined style labels,  such as speaker identities~\cite{liu2022beat}, emotions~\cite{yang2023diffusestylegesture}, or hand attributes (e.g., height, radius, and velocity)~\cite{alexanderson2020style,gesturematching22}, which are learned from annotated motion data. While effective, their performance is inherently limited by label availability and granularity, with annotation costs posing practical constraints. Example-based methods~\cite{ghorbani2022zeroeggs,aberman2020unpaired,Ao2023GestureDiffuCLIP,raab2024monkey} address this limitation by mimicking motion examples as implicit pseudo-labels~\cite{chen2024syntalker}. Although these methods achieve comprehensive control, they tend to prioritize temporally independent features and often compromise rich details present in the original motion examples.

Recent advances in large language models (LLMs) demonstrate remarkable generalization capabilities in text-related tasks~\cite{chung2024scaling}.  Through fine-tuning with structured input-output pairs, these models outperform traditional methods even in cross-modal tasks. Their universal competence has been validated in audio synthesis~\cite{fish-speech-v1.4}, robotic control~\cite{rt2}, and motion generation~\cite{jiang2024motiongpt}.

In this paper, we present \ourapproach, a framework that leverages LLMs for motion-example-controlled co-speech gesture generation. Our method capitalizes on LLMs'
comprehension capabilities through a three-stage fine-tuning mechanism to simultaneously interpret speech audio and motion examples. Departing from conventional pseudo-labeling paradigms, we position motion examples as explicit query contexts within the prompt structure to guide gesture generation, enabling the synthesis of gestures that preserve example-specific characteristics while maintaining speech congruence.

Compared with existing speech-to-gesture methods, our method achieves the state-of-the-art performance evaluated under the most used human-preference-aligned metric for gesture generation and the motion diversity metric. A user study also demonstrates that our method outperforms other example-based methods in terms of the similarity between the generated motions and the input motion examples. Furthermore, our method provides granular control of individual body parts and accommodates diverse
input modalities including motion clips, static poses, human video sequences, and textual descriptions (see Figure~\ref{fig:teaser}).

Our main contributions include:
\begin{itemize}
    \item We propose a method to directly use motion examples to control co-speech gesture generation, producing gesture motions that closely resemble the input examples.
    \item  We introduce a three-stage fine-tuning mechanism that effectively integrates audio and motion modalities into the LLM, achieving state-of-the-art performance on the speech-to-gesture task. Interestingly, this approach has a small impact on the LLM's original text comprehension capabilities.
    \item We build a comprehensive framework for co-speech gesture generation upon our example-based method, which supports multi-modal controls including motion clips, static poses, video sequences and text prompts.
\end{itemize}

\section{Related Work}

\subsection{Co-Speech Gesture Generation}

Gestures enhance the realism of artificial agents by conveying critical social cues like personality and emotional states~\cite{TheRoleofGesture}. Early gesture generation systems used rule-based methods~\cite{kopp2006bml,cassell2001beat,lee2006nonverbal,lhommet2015cerebella,cassell1994rulefullbody}, translating speech into predefined gestures via linguistic rules. However, these approaches proved labor-intensive, requiring significant manual effort for rule creation and motion segmentation. Recent advances have shifted to data-driven methods. While traditional deterministic models often produce overly smooth motions~\cite{Gesticulator,liu2022beat,yoon2020speech,zhou2022GestureMaster,gesturematching22} due to their inability to handle many-to-many mappings, modern generative models address this limitation through various architectures: normalizing flows~\cite{alexanderson2020style,ye2022audio}, VAEs~\cite{ghorbani2022zeroeggs,li2021audio2gestures,ShiCVM2024}, VQVAEs~\cite{yazdian2022gesture2vec,ao2022rhythmic,ha2g:liu2022learning,liu2022videogeneration,talkshow:yi2022generating,Lu2023CoSpeechGS}, GANs~\cite{wu2021modeling}, and diffusion models~\cite{simon2023lda,Ao2023GestureDiffuCLIP,yang2023diffusestylegesture,cheng2024siggesture,zhang2024lmm}.

\subsection{Controllable Human Motion Generation}

Speech-to-gesture mapping constitutes a many-to-many problem, where single speech signals prove inadequate for meeting users' precision demands. This necessitates integrating supplementary control signals with speech for controllable co-speech motion generation. Existing research has explored various control signals to guide gesture synthesis: motion examples~\cite{weiyu23GenMM,liu2024tango,aberman2020unpaired}, text~\cite{zhang2022motiondiffuse,tevet2023human,hong2022avatarclip,IterativeMotionEditing}, video~\cite{ha2g:liu2022learning}, images~\cite{tevet2022motionclip}, poses~\cite{ng2024audio2photoreal}, trajectories~\cite{xie2023omnicontrol,wan2023tlcontrol,karunratanakul2023gmd,shafir2024humanprior}, emotions~\cite{yang2023diffusestylegesture}, identities~\cite{liu2022beat}, hand height and radius~\cite{alexanderson2020style}.
However, such signals (e.g., emotion/identity) are typically dataset-specific and resource-intensive to acquire, limiting flexible user control.

GestureDiffuCLIP~\cite{Ao2023GestureDiffuCLIP} aligns motion sequences with CLIP embeddings~\cite{clip-pmlr-v139-radford21a} for multimodal control, yet remains constrained by CLIP's inherent motion representation limitations. To mitigate this, SynTalker~\cite{chen2024syntalker} construct a dedicated text-motion alignment space, yet still require multimodal weight balancing during inference to reconcile textual and auditory constraints. The core challenge stems from motion's inherent semantic ambiguity - while gestures can be semantically described, precise motion specification remains elusive. ZeroEGGS~\cite{ghorbani2022zeroeggs} circumvents this by directly conditioning on motion examples, but collapses arbitrary-length sequences into single-style vectors, preserving only coarse semantic attributes (e.g., emotion) while losing kinematic details. Inspired by voice cloning techniques~\cite{fish-speech-v1.4}, our approach eliminates feature extraction networks and instead directly prepends motion examples as generation prefixes, establishing explicit kinematic references for subsequent sequence synthesis.

\subsection{Multimodal LLMs}

Recent advances in large language models (LLMs) have sparked widespread multimodal extensions, with speech integration achieved by \cite{zhang2023speechgpt}, speech-image-text unification by \cite{zhan2024anygpt}, and motion incorporation through \cite{jiang2024motiongpt}. Unlike existing approaches focused on cross-modal alignment with LLM text representations~\cite{chen2024motionllm,jiang2024motiongpt,chen2024body_of_language,zhang2023speechgpt,pang2024llmgesticulatorleveraginglarge}, our framework harnesses LLMs' native capacity to decode structured inputs and approximate novel distributions. Crucially, our pipeline elimintes textual supervision beyond basic instruction formatting, revealing an emergent property: the model maintains 99\% performance parity on MMLU, GSM8K, and PIQA benchmarks compared to its original version, preserving foundational language understanding capacities.

Our work is also related to recent methods utilizing language models to synthesize motions. T2M-GPT~\cite{zhang2023generating} employs the GPT architecture to perform text-to-motion tasks. MotionGPT~\cite{jiang2024motiongpt} finetunes T5~\cite{2020t5} to tackle various text-motion tasks. M$^3$GPT~\cite{luo2024m3gpt} further extends it by incorporating text-music-dance related tasks. None of these methods is designed for the co-speech gesture generation task, and it is difficult to conduct direct comparisons with these methods. We discuss them in more details in Section 3 of the supplementary material.

\section{Method}

\begin{figure}[t]
  \centering
  \includegraphics[width =\linewidth]{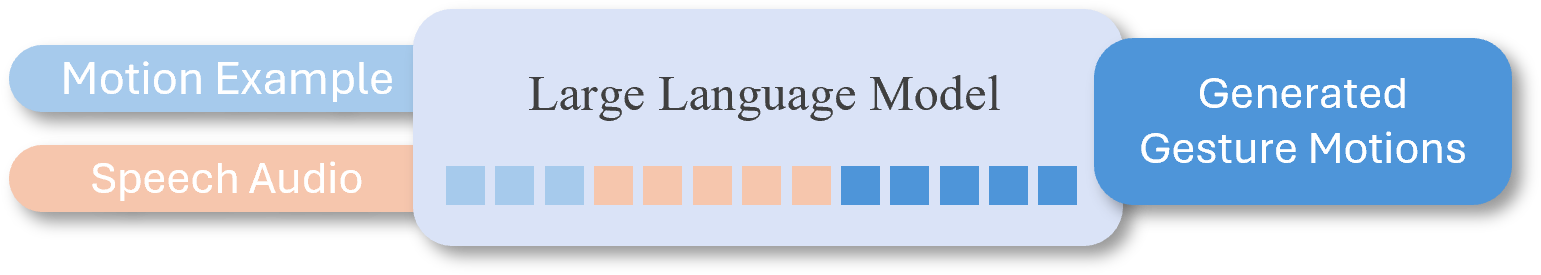}
  \caption{Our model takes motion examples and speech audio as inputs. Both inputs are converted into token sequences by tokenizers and fed into an LLM for autoregressive generation. The generated motion tokens are then processed through a motion decoder to produce the target gesture motion.}
  \label{fig:2model}
\end{figure}

Given a speech audio and a reference motion sequence, our goal is to synthesize co-speech gestures with stylistic consistency to the reference motions, as depicted in Figure~\ref{fig:2model}. By harnessing LLMs' dual capabilities in instruction following and conditional generation, we develop a multimodal fusion framework that processes both auditory and kinematic inputs to produce contextually appropriate co-speech gesture motions.

\begin{figure*}[t]
  \centering
  \includegraphics[width = 0.9\linewidth]{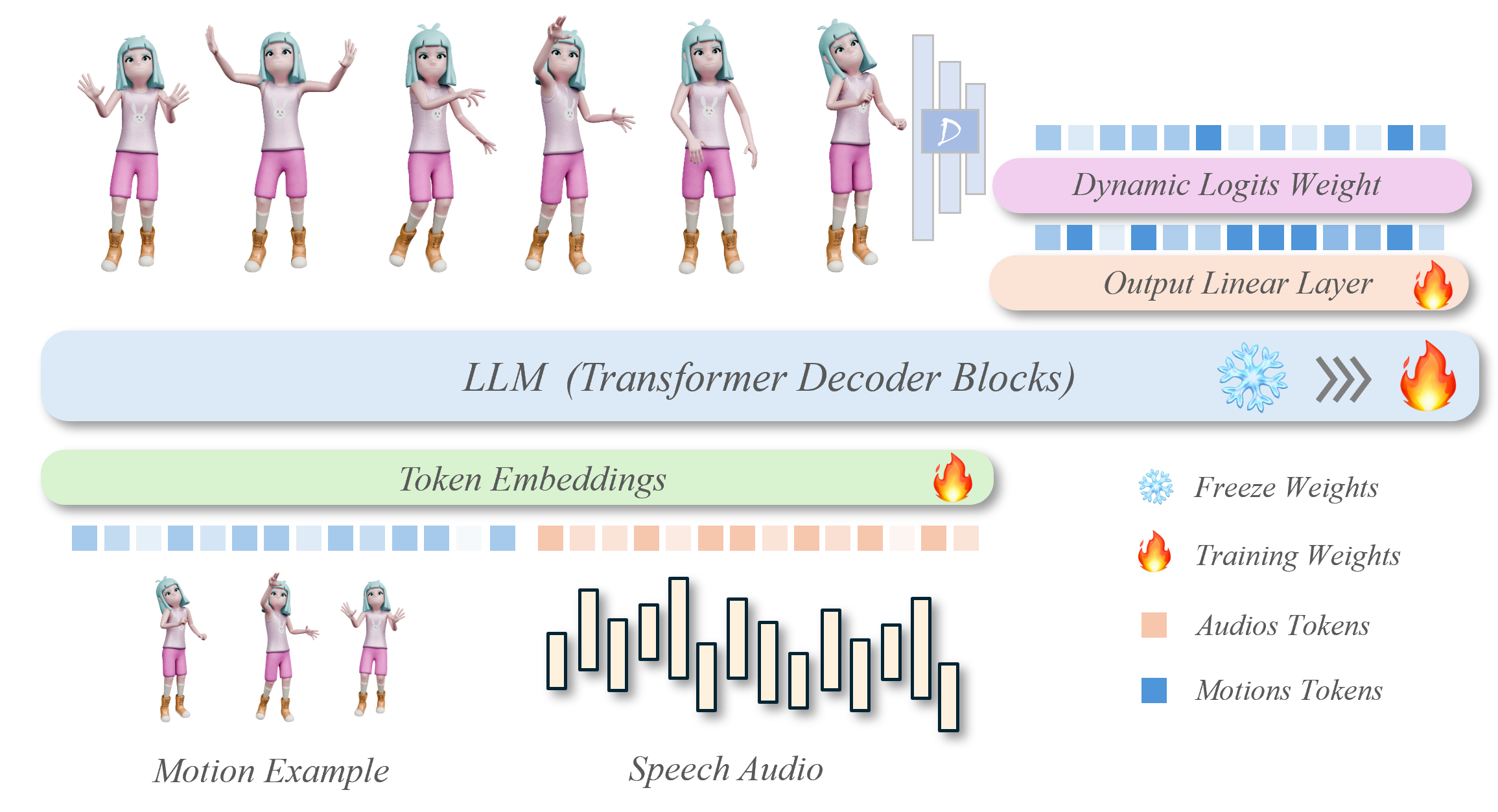}\vspace*{-3mm}
  \caption{The structure of our example-guided co-speech generation model. Both motion and audio are tokenized and fed into a large language model (LLM) to generate co-speech motion tokens. Initially, we fine-tune the embedding layer and output linear layer (unembedding space) to adapt the new tokens to the token distribution of the LLM. Subsequently, we perform full parameter fine-tuning to enable the LLM to generate motion tokens.}
  \label{fig:model}
\end{figure*}

To enable LLMs to comprehend speech audio and motion data, these multimodal inputs must first be mapped to tokens within the LLM's embedding space. However, directly training models with randomly initialized tokens presents significant challenges, as their initial distribution diverges from the pre-existing token embedding distribution of the base LLM. This misalignment causes early-stage training instability and hinders effective utilization of the LLM's inherently well-structured parameter space, which risks degrading the model's original capabilities. To address this issue, we propose a novel token initialization method. As shown in Figure~\ref{fig:model}, during initialization, only the parameters associated with the newly introduced tokens are made trainable. This strategy yields more optimal initial values for the additional tokens, ensuring enhanced compatibility with the established embedding space while preserving the integrity of the pre-trained model.

Building upon this initialization, we employ two training stages to enable example-controlled co-speech gesture generation. The first stage exclusively trains the model's speech-to-gesture mapping capability, establishing core correlations between these two modalities. The subsequent stage introduces motion-example-conditioned training objectives, where the model learns to adapt gestures to both speech content and reference motion examples. This progressive training strategy significantly enhances generation robustness, particularly in data-scarce scenarios where available motion examples are limited and insufficient for gesture generation.

To enhance practical applicability, we design a parameterized sampling mechanism during inference that provides users with granular controllability over motion-example adherence levels. This continuous spectrum ranges from strict compliance with reference motions to partial integration or complete ignorance, enabling context-aware adaptation of gesture generation fidelity. We further generalize the framework to incorporate various input modalities, including poses, video sequences, and textual descriptions, for more flexiable control of gesture generation.

\subsection{Motion Representation}
\label{subsec:motion_representation}

A motion $\mathbf{m}_{1:N}\in\mathbb{R}^{N\times  (4+6J)}$ is a sequence of poses, where $N$ denotes the motion length. Each pose $m \in \mathbb{R}^{4+6J}$ consists of root angular velocity along Y-axis, 
root linear velocities on XZ-plane, root height and the rotations of its $J$ joints, where the rotations are parameterized as 6D vectors~\cite{Zhou_2019_CVPR}.

For simplicity, the motion sequence is represented as $\mathbf{m}_{1:N}\in\mathbb{R}^{N\times{(4+6J)}}$. It is firstly encoded into a latent vector sequence $\mathbf{{z}}_{1:n}\in\mathbb{R}^{n\times f}$ with a downsampling ratio of $n/N$ and latent feature dimension $f$, using 1D convolutional encoder $\mathcal{E}$.
\cbhb{The $\mathbf{{z}}_{1:n}\in\mathbb{R}^{n\times f}$ obtained through the encoder then enters the base quantization layer $\mathrm{Q}_0$.} Each vector subsequently finds its nearest code entry in the layer's codebook $\mathbf{C}_0=\{\mathbf{c}^0_k\}_{k=1}^{K}\subset \mathbb{R}^f$ to get the quantization vector \cbhb{${\hat{\mathbf{z}}}^{0}_{1:n}$}. We calculate the quantization residual \cbhb{${\mathbf{r}}_{1:n} = {\hat{\mathbf{z}}^0_{1:n}} - \mathbf{{z}}_{1:n}$}, which then enters the first residual quantization layer $\mathbf{Q}_1$ and finds its nearest code entry in the layer's codebook $\mathbf{C}_1=\{{\mathbf{c}^1_k\}}_{k=1}^{K}\subset \mathbb{R}^f$ to get the first residual quantization vector \cbhb{${\hat{\mathbf{z}}^1_{1:n}}$}. Accordingly, \cbhb{${\hat{\mathbf{z}}^2_{1:n}}$,${\hat{\mathbf{z}}^3_{1:n}}$}... can be calculated in this manner. 
As the final step of motion encoding, we sum all quantization vectors together to get the final code $\hat{\mathbf{z}} = \sum_{q=0}^Q {\hat{\mathbf{z}}}^q_{1:n}$, where $q=0$ corresponds to the base quantization layer and $q \in\{1,2,...,Q\}$ represent the residual quantization layers. Then $\hat{\mathbf{z}}$ is fed into decoder $\mathcal{D}$ for decoding it to motion ${\hat{\mathbf{m}}_{1:N}}$.

To train the encoder/decoder and all codebooks, we execute the reconstruction task using the following loss function
\begin{align}
    \cbhb{\mathcal{L}_{rec} = \|{{\hat{\mathbf{m}}_{1:N}}}-{\mathbf{m}_{1:N}}\|_1 + 
    \eta\sum_{q=0}^Q\|\mathbf{z}^{q}_{1:n}-\mathrm{sg}[{\hat{\mathbf{z}}}^{q}_{1:n}]\|_2^2,}
\end{align}
where $\mathrm{sg}[\cdot]$ denotes the stop-gradient operation, and $\eta$ is a weighting factor for the embedding constraint.

To maximize the information captured in the first quantization layer, we randomly drop subsequent residual quantization layers after the base quantization layer during model training~\cite{guo2023momask}. This ensures that the base quantization layer learns to encode as much information as possible. We only utilize the base quantization layer without performing VQ completion operations like~\cite{Zhang2024SemanticGesture,guo2023momask}. Although our approach shares the same architecture with vanilla VQVAE during inference, it achieves superior performance in both reconstruction quality and downstream tasks (see detailed validations in Section~2 of the supplementary material).

Given our objective to enable motion-example-guided gesture synthesis, we confront practical constraints in obtaining comprehensive full-body motion examples. This is particularly relevant as many current motion-related works primarily focus on either upper body movements or full-body motions excluding fingers. To bridge this gap, we implement anatomically partitioned tokenization through three functional regions: upper body, lower body, and hands. Each partition is encoded and trained separately, using the same training process. During synthesis, all body partitions are generated simultaneously (see implementation details in Section~1 of the supplementary material). For description simplicity, we use full-body motion modeling as the baseline configuration in the following.

\subsection{Finetune LLM}
\subsubsection{Token embedding initialization}
\label{subsubsec_stage1}
We use the encoder $\mathcal{E}$ to tokenize motion sequence $\mathbf{m}_{1:N}$ into a sequence of discrete units $\mathbf{c}_{1:T_c}$  and a Hidden-unit BERT (HuBERT)~\cite{hsu2021hubert} to encode speech audio $\mathcal{A}$ into a sequence of discrete units ${\mathbf{a}_{1:T_a}}$. Since the origin LLM does not have corresponding audio and motion tokens, we need to first extend the vocabulary of the LLM. 
Token embedding is crucial in LLMs, especially for tokens representing new modalities. To further find better initialization values for these new tokens that leverage the LLM's existing capabilities while minimizing disruption to its core functions, we initially freeze the main LLM parameters and only train the token embedding layer and the final output projection layer using the LLM's original pretraining task
\begin{align}
\cbhb{\mathcal{L}(\theta_0) = -\sum_{t=1}^{T_a} \log p(\mathbf{a}_t \mid \mathbf{a}_1, \mathbf{a_2}, \dots, \mathbf{a}_{t-1})\nonumber} \\
\cbhb{-\sum_{t=1}^{T_c} \log p(\mathbf{c}_t \mid \mathbf{c}_1, \mathbf{c}_2, \dots, \mathbf{c}_{t-1}),}
\end{align}
where \(\theta_0 = (\theta_{\text{embedding}}, \theta_{\text{output projection}})\).
\subsubsection{Speech to gesture}
\label{subsubsec_stage2}

Building upon the previous step, we enable training of all LLM parameters while fine-tuning the model by executing Supervised Fine-Tuning (SFT)~\cite{ouyang2022instructgpt} tasks
\begin{align}
\mathcal{L}(\theta) = -\sum_{t=1}^{T_c} \log p(m_t \mid \mathbf{a}_1, \mathbf{a}_2, \dots, \cbhb{\mathbf{a}_{T_a}, \mathbf{c}_1, \mathbf{c}_2, \dots, \mathbf{c}_{t-1})}.
\end{align}

In this training setup, speech audio serves as the user's query in the conversation, while gesture motion acts as the assistant's response. This structure naturally frames our task within the typical conversation format used in LLM training. Note the main purpose of this training step is to establish a mapping between the two modalities in the LLM (audio and motion).

\subsubsection{Example-controlled co-speech gesture generation}
\label{subsubsec_stage3}
Finally, we further finetune the LLM augmented with motion examples. The target generated gesture token sequence naturally serves as a motion example input. However, using it directly as a condition which would lead to the generated motions directly copying the motion example while ignoring the audio. Therefore, we process the token sequence through deduplication and shuffling operations to create our motion example. Deduplication removes repeated tokens from the token sequence, while shuffling randomly reorders the tokens in it.

Additionally, since in practice, the motion examples provided by users may be insufficient to generate reasonable co-speech gestures, we also want the model to automatically fill in missing movements during the process. Therefore, we incorporated a random dropout operation on the motion example elements during training. This process can be described as 
\begin{align}
{\mathbf{E}_c}=\text{Drop\&Shuffle\&Dedup}(\cbhb{\mathbf{c}_1, \mathbf{c}_2, \dots, \mathbf{c}_{T_c}}).
\end{align}

In this stage, we train using the same data as in the previous stage while using the  following loss function
\begin{align}
\mathcal{L}(\theta) &= -\sum_{t=1}^{T_c} \log p(\mathbf{c}_t \mid \mathbf{E}_c, \mathbf{a}_1, \mathbf{a}_2, \dots, \cbhb{\mathbf{a}_{T_a}, \mathbf{c}_1, \mathbf{c}_2, \dots, \mathbf{c}_{t-1}})\nonumber \\
&\quad + \lambda \sum_{i=1}^{T_c} p(\cbhb{\mathbf{c}_i \notin \{ \mathbf{c}_1, \mathbf{c}_2, \dots, \mathbf{c}_{T_C}} \}).
\end{align}

We further utilize the motion example ${\mathbf{E}_c}$ as a system prompt to assist in the generation process. To ensure the model effectively learns from the given motion examples, we introduce an additional penalty term to the loss function, which discourages the model from generating outputs that deviate from the motion examples. In details, We check the probability of tokens not appearing in motion example and punish their probability. It is important to note that the motion examples used in this penalty term are the original data without dropout operations.

\subsection{Inference}
\label{sec:3.3_inference}

Similar to traditional text-based LLMs, during model inference, we use audio as the user query and motion examples as system prompts. To accommodate the requirement of specific initial character states, we set the character's initial pose as the starting point of the model's answer sequence. For arbitrary long audio sequences, we adopt a segmented generation approach, with each segment having the same length as the audio used during training. To ensure temporal consistency across each generated motion segment, when generating the next segment, we use the last three codes of the currently generated motion as the answer sequence, with the corresponding audio aligned accordingly.

To control the frequency of example motions in the generated sequence, we propose a logit-based sampling strategy. Instead of merely adjusting the sampling temperature, which could lead to motion discontinuities, we introduce a manually selected hyperparameter, denoted as \(\beta\), to the logits of tokens corresponding to motion examples. Additionally, to avoid the repetitive sampling of specific tokens and promote diversity in the motion examples, we apply a decay factor, \(\gamma\), to the logits of each subsequent token after one is sampled. The adjusted logits are given by
\begin{align}
{logits}_i' &= ({logits_i} + \beta) \cdot \gamma^t,
\end{align}
where \(i\) denotes the index of the \(i\)-th token, and \(t\) represents the frequency of occurrence of the \(i\)-th token in the previously sampled sequence. The computed \({logits}_i'\) replaces the original logit in the subsequent softmax computation to derive the final sampling probability.

\section{Experiments}
\subsection{System Setup}

\subsubsection{Dataset}
We train and test our model on two high-quality mocap co-speech gesture datasets: 
(1) BEAT2~\cite{liu2024emage} provides 60 hours of SMPL-X~\cite{SMPL-X:2019}  formatted full-body motion from 25 speakers. Following the benchmark protocol, we only use the second speaker's data for training and testing; (2) ZeroEGGS~\cite{ghorbani2022zeroeggs} features two hours of English monologue data from a female performer across 19 different styles, including synchronized motion and audio. We use the same datasets split as in their work.

\subsubsection{Settings}

Our system generates motions at 30 frames per second. The motion RQVAEs, described in Section ~\ref{subsec:motion_representation}, are trained with a downsampling ratio of ${n/N}$ = 4, K = 512, d = 512, Q = 6, batch size = 256, \cbhb{$\eta$ = 0.1}, a learning rate of 4e-4, and a step learning rate scheduler. 
For fine-tuning the LLM, we use Qwen2.5-0.5b-instruction~\cite{qwen2.5} as our base model and detach its tied embeddings. In Sections ~\ref{subsubsec_stage1}, ~\ref{subsubsec_stage2}, and ~\ref{subsubsec_stage3}, the batch sizes per GPU are 32, 20, and 12, respectively. The gradient accumulation steps are set to 4, 6, and 10, and the learning rates are 2e-4, 5e-5, and 5e-5 for each stage. We use the cosine schedule with warmup as our learning rate scheduler. 
In Sections ~\ref{sec:3.3_inference}, $\beta$ is set to 5 by default, and $\gamma$ = 0.9.

During RVQVAEs training, we randomly sample 64-frame motion sequences. For LLM fine-tuning, given Hubert's audio encoding rate of 50Hz and motion encoding rate of 7.5Hz, we use 4-second audio and motions, corresponding to 200 audio tokens and 90 motion tokens, with 30 tokens for each of the three body parts. We train all these models using four NVIDIA RTX 4090 GPUs in 22 hours. During inference, using an NVIDIA RTX 4090 GPU and Intel i9-13900KF CPU, our model achieves an inference speed of 147 tokens per second with the default Hugging Face inference pipeline~\cite{wolf-etal-2020-transformers}. When deploying the same model using the vLLM~\cite{kwon2023efficient} inference framework, the generation throughput increases to 270 tokens per second. It means that we can generate 36 seconds of motion in just 1 second.

\begin{table}[t]
\centering
\caption{Comparison with the state-of-the art methods on BEAT2~\cite{liu2024emage} test set. Quantitative evaluation on BEAT2. We report FGD $\times 10^{-1}$, BC $\times 10^{-1}$, and diversity. \textbf{Bold} face indicates the best result.}
{
\begin{tabular}{lccc}
\toprule
     Method & FGD $\downarrow$ & BC $\uparrow$ & Diversity~$\uparrow$   \\ 

\midrule
S2G\cite{ginosar2019learning} & 28.15  & 4.683  & 5.971                        \\
Trimodal\cite{yoon2020speech} & 12.41  & 5.933  & 7.724                         \\
HA2G\cite{ha2g:liu2022learning} & 12.32  & 6.779 & 8.626                        \\
DisCo\cite{liu2022disco} & 9.417  & 6.439 &  9.912                              \\
CaMN\cite{liu2022beat} & 6.644  & 6.769 & 10.86                                  \\
DiffStyleGesture\cite{yang2023diffusestylegesture} & 8.811 & 7.241 & 11.49     \\
Habibie \textit{et al}.\cite{habibie2021learning} & 9.040 &  7.716 &  8.213       \\
TalkShow\cite{talkshow:yi2022generating} &  6.209 &  6.947 & 13.47                \\      
SynTalker\cite{chen2024syntalker}&  6.413  & \textbf{7.971} & 12.72                      \\ 
EMAGE \cite{liu2024emage} & 5.512 &  7.724 & 13.06                               \\

\midrule

\ourapproach~ & 3.401 & 7.346  & 15.30 \\
\ourapproach~(w/ examples) & \textbf{2.999} & 7.472  & 15.01 \\
\ourapproach~(7b llm) & 3.456 & 7.470  & \textbf{15.64} \\
\ourapproach~(7b llm w/ examples) & 3.195 & 7.554  & 15.46 \\
\midrule
\ourapproach~(w/o freeze) & 8.512 & 4.551  & 13.46 \\
\ourapproach~(w/o freeze\&pretrain) & 4.575 & 6.936 & 15.45 \\
\ourapproach~(w/o s2g) & 4.845 & 6.910  & 15.09 \\
\ourapproach~(w/o s2g ; w/ examples) & 4.413 & 7.138  & 14.76 \\
\ourapproach~(w/o llm) & 10.32 & 5.813 & 13.47  \\
\ourapproach~(w/o Instruct llm) & 4.133 & 6.962 & 15.13  \\
\bottomrule
\end{tabular}}
\label{table1}
\end{table}

\begin{table*}[t]
\caption{Comparison of the similarity between the generated results and the motion example. The values in this table represent the mean and standard deviation, where the standard deviation is shown after '±'.}
\centering
\begin{tabular}{llcccc}
\toprule
Dataset & System & $\text{FGD1}_{\text{train}}\downarrow$ & $\text{FGD1}_{\text{test}}\downarrow$ & $\text{FGD2}_{\text{train}}\downarrow$ & $\text{FGD2}_{\text{test}}\downarrow$\\
\midrule
\multirow{5}{*}{ZEGGS} 
& ZeroEGGS &  $3.39\pm 0.179 $ & $4.54\pm 0.282 $ & $23.14\pm 1.973 $ & $26.04 \pm 1.939$\\
& \ourapproach & $1.83\pm 0.118 $ & $1.98 \pm 0.593 $ & $10.22 \pm 2.221$ & $10.20 \pm 2.881$\\
\cline{2-6}

& \ourapproach ~ (w/o freeze\&pretrain) & $2.47\pm 0.239 $ & $2.82 \pm 0.647 $ & $16.63 \pm 2.752$ & $18.94 \pm 3.326$\\
& \ourapproach ~ (w/o s2g) & $2.29\pm 0.137 $ & $2.65 \pm 0.683 $ & $13.47 \pm 2.390$ & $14.98 \pm 3.217$\\
& \ourapproach ~ (w/o instruct llm) & $1.96\pm 0.153 $ & $2.31 \pm 0.591 $ & $11.86 \pm 2.528$ & $12.14 \pm 2.946$\\

\midrule
\multirow{5}{*}{BEAT2} 

& SynTalker & $4.19 \pm 0.477$ & $8.21 \pm 0.771 $ & $24.74 \pm 1.455$ & $37.72 \pm 3.136$\\

& \ourapproach & $2.65 \pm 0.243$ & $4.12\pm 0.513 $ & $17.23 \pm 1.338$ & $21.73 \pm 2.647$\\
\cline{2-6}
& \ourapproach ~ (w/o freeze\&pretrain) & $2.95\pm 0.206 $ & $4.55 \pm 0.483 $ & $20.41 \pm 1.452$ & $26.17 \pm 2.441$\\
& \ourapproach ~ (w/o s2g) & $2.83\pm 0.272 $ & $4.68 \pm 0.656 $ & $19.11 \pm 1.539$ & $28.53 \pm 2.968$\\
& \ourapproach ~ (w/o instruct llm) & $2.81\pm 0.254 $ & $4.29 \pm 0.577 $ & $18.62 \pm 1.430 $ & $22.95 \pm 2.703$\\
\bottomrule
\end{tabular}
\label{table2}
\end{table*}

\subsection{Comparisons}

\subsubsection{Speech-to-gesture benchmark}

In the traditional speech-to-gesture task, in order to avoid leaking information from the test dataset, the motion examples in the model's input conditions are set to be empty during sampling. To ensure the reproducibility of results, only greedy sampling is used when calculating quantitative metrics. We have achieved SOTA performance as shown in Table \ref{table1}, particularly on FGD~\cite{yoon2020speech}, which is currently the most used human-preference-aligned metric for gesture generation evaluation~\cite{genea2022}. We further test the inference process with the inclusion of motion examples (w/ examples), and find that our results are further improved. These results not only demonstrate our model's superiority in the pure speech-to-gesture task but also validate the effectiveness of motion examples in controlling co-speech gesture generation.

\subsubsection{Speech-to-gesture with motion examples}

3
For the co-speech gesture generation task, we select two works that similarly support motion examples as input for comparison: ZeroEGGS~\cite{ghorbani2022zeroeggs} and SynTalker~\cite{chen2024syntalker}. While ZeroEGGS is originally trained on the ZeroEGGS dataset, we follow its data split to retrain our model on it. For SynTalker, we directly use their published code and pre-trained checkpoints.

\begin{figure*}[t]
  \centering
  \includegraphics[width = \linewidth]{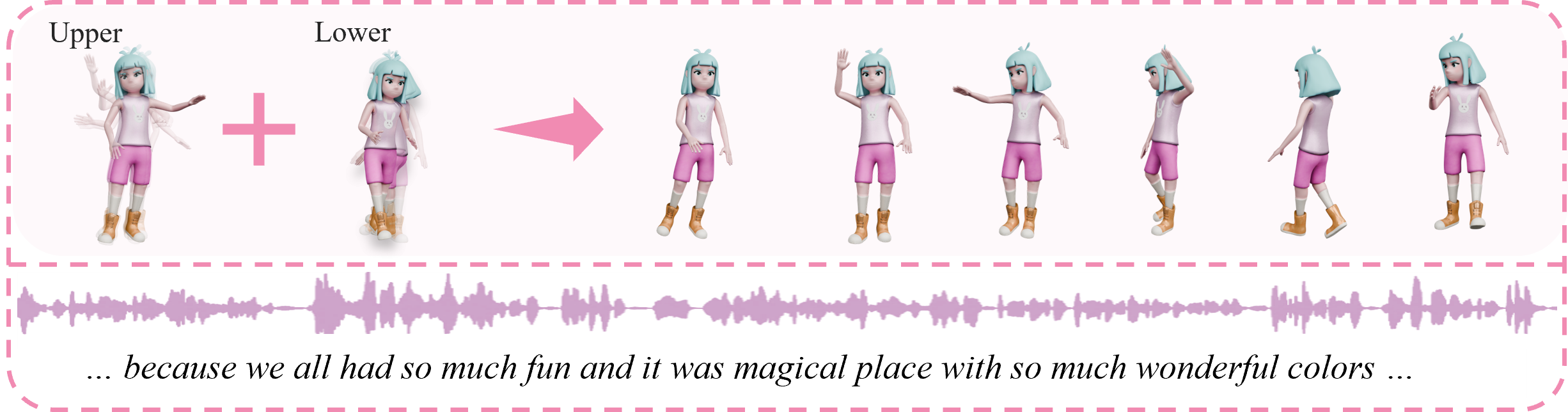}
  \caption{We can control specific body parts by tokenizing examples and combining their corresponding tokens. For instance, we tokenize two examples, use the upper body token from the first and the lower body token from the second as a prompt. The generated motion effectively reflects these references.}
  \label{fig:part_control}
\end{figure*}

\begin{figure*}[t]
  \centering
  \includegraphics[width = \linewidth]{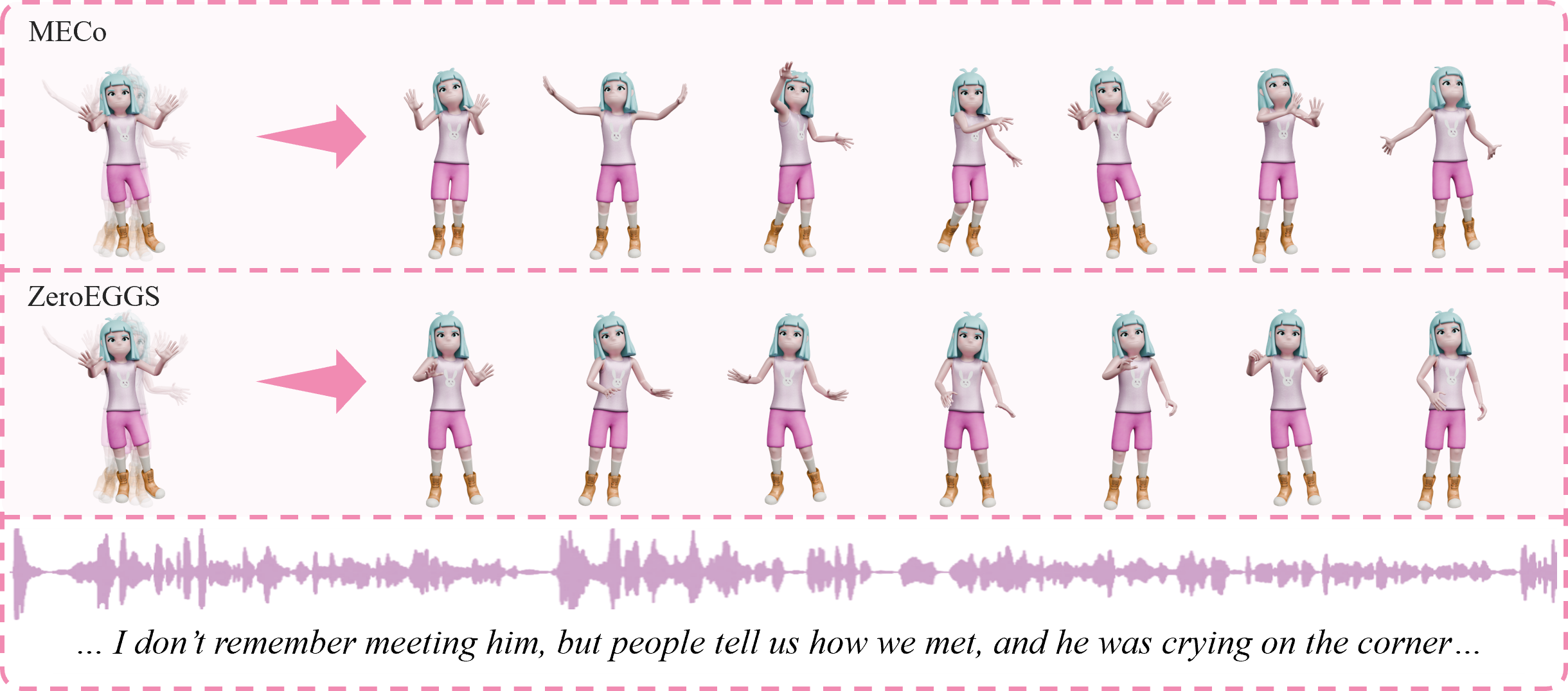}
  \caption{A qualitative comparison between our method and ZeroEGGS. Both methods use the same input, with the motion example displayed on the left side of the arrows in the figure, and the input audio presented at the bottom of the figure.}
  \label{fig:cmp_zeroeggs}
\end{figure*}

\begin{figure*}[t]
  \centering
  \includegraphics[width = \linewidth]{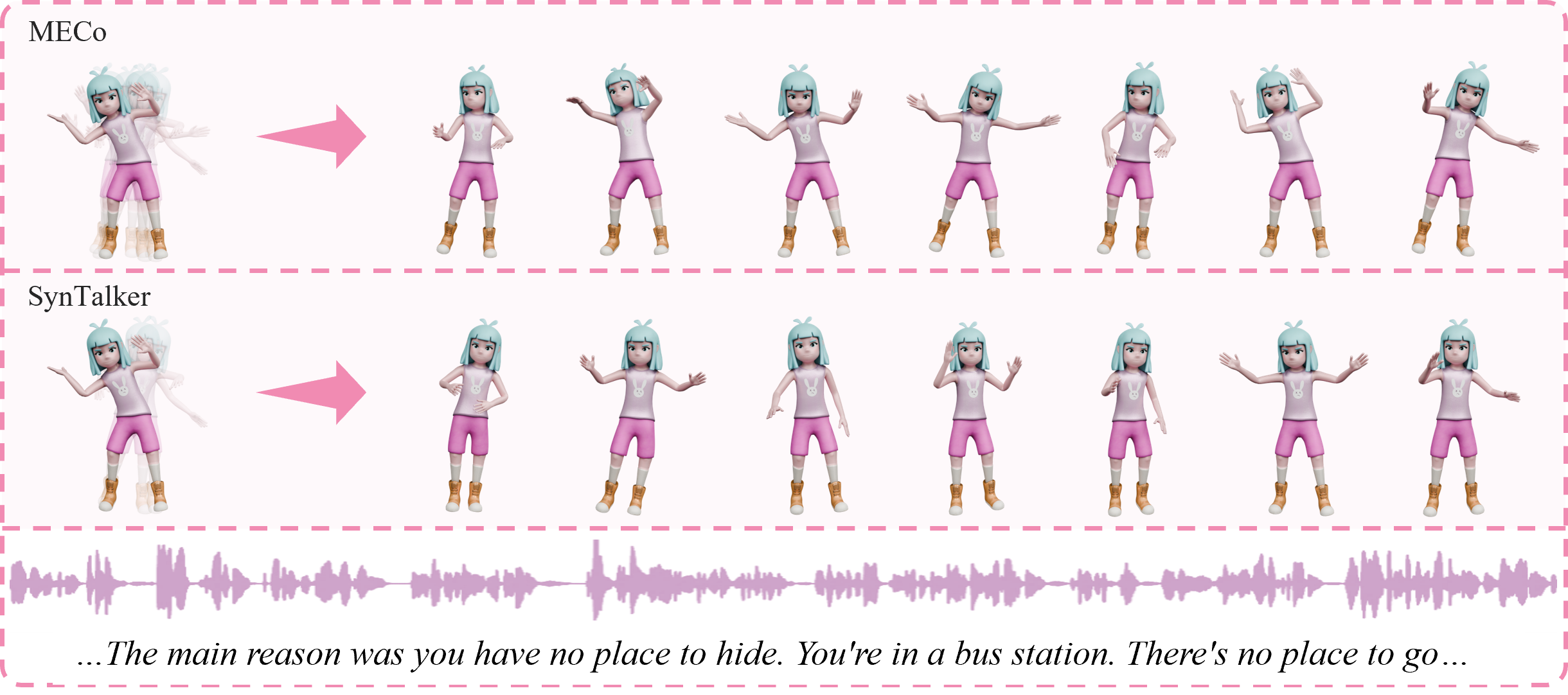}
  \caption{A qualitative comparison between our method and SynTalker. Both methods use the same input, with the motion example displayed on the left side of the arrows in the figure, and the input audio presented at the bottom of the figure.}
  \label{fig:cmp_syntalker}
\end{figure*}

To validate whether the generated motions follow the examples, we still use FGD as the metric, as it is the most effective in determining whether two sequences are similar. Since the ZeroEGGS dataset does not provide a feature extractor, following ~\cite{yoon2020speech,liu2022beat}, we use an autoencoder as our feature extractor to compute the Fr\'echet distance, which we refer to as FGD1. Since this autoencoder is trained by us, for fairness, like ~\cite{ng2023text2listen}, we directly calculate the Fr\'echet distance on the representations of the motions themselves, which we refer to as FGD2. For the BEAT2 dataset, we compute FGD1 using its default feature extractor~\cite{skeletoncnn}. To compare the performance of different methods on both training and test sets, we sampled 20 motion sequences with durations ranging from 3 to 6 seconds as motion examples. For input audio, we consistently used a neutral-style audio clip from the test set. For each generated result, we computed its FGD1 and FGD2 with the corresponding motion example. The final metrics are calculated by averaging all results. The experimental results in Table~\ref{table2} show that our method achieves superior performance compared to existing approaches on both training and test sets. Figure~\ref{fig:part_control} demonstrates that we can use different motion examples for different body parts to achieve granular control. Additionally, the visualization comparison results in Figure ~\ref{fig:cmp_zeroeggs} and ~\ref{fig:cmp_syntalker} clearly show that our generated results are more closely aligned with those of the motion example.

\subsection{User Study}
Following ~\cite{simon2023lda, Ao2023GestureDiffuCLIP, Zhang2024SemanticGesture}, we conduct a similar user study to validate the effectiveness of our method. For each method to be evaluated, eighty 10-second audio segments were employed to generate animations. 29 participants were recruited for this study, and each questionnaire included 24 paired comparisons. They selected their preferred clip and rated their preference intensity on a 0-2 scale (0 indicating no preference). The unselected clip automatically received the opposite/negative score. We evaluate the generated gestures using three subjective metrics: \textit{Human Likeness} for realism and human-like quality, \textit{Appropriateness} for alignment with speech rhythm and semantics, and \textit{Example Consistency} for similarity to the reference motion example. As shown in Table~\ref{table:user_study}, our method achieves the best performance under these metrics, especially in the \textit{Example Consistency}.

\subsection{Ablation Study}

\subsubsection{Initialize token embedding}

To validate the importance of this step in our experiment, we conducted two ablation studies. We first directly omit our embedding initialization step described in Section~\ref{subsubsec_stage1}\, and use the default token embedding initialization in PyTorch for the newly added vocabulary. 

As shown in Tables~\ref{table1} and Table~\ref{table3}, this method led to a degradation in co-speech gesture generation quality metrics. Moreover, it impairs some of the LLM's inherent capabilities. We also provide more detailed comparisons of models in terms of degradation rates in Section~5 of the supplementary material.

We further conduct an ablation study on the partial freezing strategy described in Section~\ref{subsubsec_stage1}, where we make the main body of the LLM frozen. In this experiment, we made all model parameters trainable during the training process, instead of our original approach where only token embeddings and the output linear layer are trainable. As shown in Tables ~\ref{table1} and ~\ref{table3}, this modification severely degrades the model's performance, resulting in the generation of static motions lacking any meaningful variation.

\subsubsection{Speech-to-gesture Training}

In this experiment, we investigate eliminating the speech-to-gesture training phase prior to example-controlled co-speech gesture training (i.e., using only Sections~\ref{subsubsec_stage1} and~\ref{subsubsec_stage3}). This decision is motivated by the valid concern that Section~\ref{subsubsec_stage3} already involves training the speech-to-gesture task, raising the question of whether a dedicated training phase is necessary. As shown in Table~\ref{table1}, after removing this training task, the performance metrics for co-speech gesture generation show a noticeable decline for both \cbhb{MECo (w/o s2g) and (w/o s2g; w/ examples)}. Here, ``w/o s2g'' refers to training without the speech-to-gesture task, as described in Section~\ref{subsubsec_stage2}, and ``w/ examples'' indicates the use of motion examples during the metric evaluation. Including the speech-to-gesture task during training helps improve the quality of gesture generation, especially in cases where no motion examples are given or the provided motion examples are short in length.

\begin{table}[t]
\caption{Comparison with the original LLM in performance metrics.}
\begin{tabular}{lccc}
\toprule
{Model} & {MMLU}$\uparrow$ & {GSM8K}$\uparrow$ & {PIQA}$\uparrow$\\ 
\midrule
Qwen2.5-0.5b-instrcut & 46.50 & 20.47  & 70.13\\ 
\ourapproach~ & 46.27 & 20.47  & 69.64\\ 
\ourapproach~ (w/o freeze) & 24.63 & 0.15 & 52.50\\ 
\ourapproach~ (w/o freeze\&pretrain) & 39.62 & 15.24 & 65.83\\ 
\midrule
Qwen2.5-7b-instrcut & 74.20 & 82.18 &  80.30\\ 
\ourapproach~ (7b llm) & 74.13 & 81.96 &  79.54\\ 

\bottomrule
\end{tabular}
\label{table3}
\end{table}

\subsubsection{LLM backbone}
In this experiment, to validate the effectiveness of instruction LLM for our work, we introduced two additional LLM variants for comparison: a base LLM that only underwent pre-training, and an untrained LLM with randomly initialized parameters. It is worth noting that all LLMs share the same network architectures, differing only in their parameter values. As demonstrated in Table~\ref{table1} and Table~\ref{table2}, the instruction-tuned LLM consistently outperforms the other two variants in gesture generation capabilities in both co-speech gesture generation ability and motion examples followed ability.

\begin{figure*}[t]
  \centering
  \includegraphics[width = \linewidth]{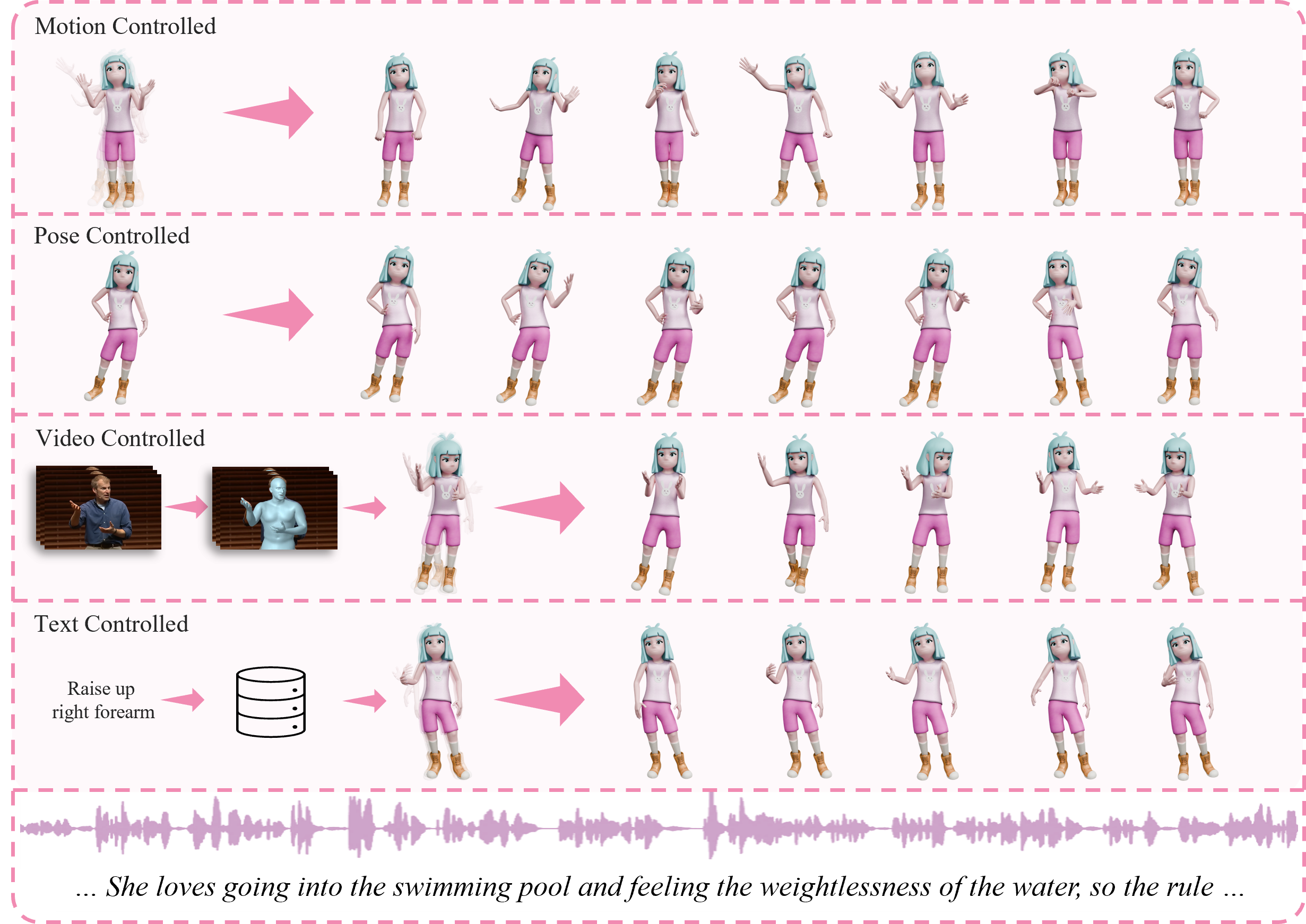}
  \caption{We demonstrate the versatility of our method across various control modalities, including direct motion control, pose control, video control, and text control. These diverse modalities are unified as motion examples, which serve as prompts for our system. By leveraging these prompts, our method performs co-speech motion generation that is not only aware of the speech audio but also aligned with the provided motion examples to reflect user intent.}
  \label{fig:4_controls}
\end{figure*}

We also conducted parameter scaling tests using Qwen2.5-Instruct (7B parameters). As shown in Table~\ref{table1}, no scaling benefits were observed - performance actually slightly decreased. This likely indicates that current co-speech gesture data scarcity renders even 0.5B models sufficiently capable, leaving no advantage for larger architectures. For instance, the current BEAT2 benchmark uses the second character's data (containing only about 2 hours of co-speech gestures) as its test set. After tokenization, this yields only 54k motion tokens and 360k audio tokens. Even with data augmentation through mirroring and speed variation, this scale remains minimal compared to text models like the 0.5B-parameter model in Qwen2.5~\cite{qwen2.5}, which was trained from scratch on 18T tokens.

\begin{table*}
\caption{User study of different systems on BEAT2 and ZeroEGGS datasets. The results are reported as average scores with 95\% confidence intervals.}
\centering
\begin{tabular}{llccc}
\toprule
Dataset & System & HumanLikeness $\uparrow$ & Appropriateness $\uparrow$ & ExampleConsistency $\uparrow$ \\
\midrule
\multirow{5}{*}{BEAT2} 
& EMAGE & $-0.59 \pm 0.17$ & $-0.48 \pm 0.19$ & $-$ \\
& SynTalker(U) & $0.18 \pm 0.19$ & $0.12 \pm 0.19$ & $-$ \\
& SynTalker & $-0.28 \pm 0.20$ & $-0.42 \pm 0.22$ & $-0.64 \pm 0.21$ \\
& \ourapproach(U) & $0.34 \pm 0.18$ & $0.30 \pm 0.20$ & $-$ \\
& \ourapproach & $0.28 \pm 0.20$ & $0.42 \pm 0.22$ & $0.64 \pm 0.21$ \\
\midrule
\multirow{2}{*}{ZeroEGGS} 
& ZeroEGGS & $-0.53 \pm 0.24$ & $-1.10 \pm 0.16$ & $-1.17 \pm 0.17$ \\
& \ourapproach & $0.53 \pm 0.24$ & $1.10 \pm 0.16$ & $1.17 \pm 0.17$ \\
\bottomrule
\end{tabular}
\label{table:user_study}
\end{table*}

\subsection{Multimodal Controls}

Our example-based method can support diverse input modalities. For video input, we extract SMPL-X parameters via monocular motion capture~\cite{talkshow:yi2022generating}.
For image input, we reconstruct static poses using SMPLify-X~\cite{SMPL-X:2019}. For text input, we employ two approaches: 1) Annotate our co-speech gesture dataset with text labels to establish gesture-text mappings; 2) Train a motion-text retrieval system (TMR~\cite{petrovich23tmr}). Given text queries, ChatGPT first checks for matching annotations in our gesture dataset. If available, the corresponding motion is used; otherwise, TMR retrieves the most similar motion from the broader corpus. Figure~\ref{fig:4_controls} demonstrates our framework’s multimodal operation and outputs.

\section{Discussions and Future Work}
Our experiments reveal degraded performance when using video prompts. Analysis shows that for many in-the-wild videos, monocular motion capture derived SMPL-X parameters exhibit significant reconstruction errors after VQ-VAE processing. This highlights our motion VQ-VAE's limited generalization to out-of-distribution data. Addressing this limitation by improving the VQ-VAE's generalization capability constitutes our next research priority.

Our generated motions also exhibit physically implausible artifacts (e.g., foot sliding), a common challenge in kinematic motion generation, which could be addressed via inverse kinematics (IK) or physics-based simulations~\cite{Luo2023PerpetualHC,MoConVQ}.

\begin{acks}
This work is partially supported by NSF China (No. 62421003,  62206245) and the XPLORER PRIZE.
\end{acks}

\bibliographystyle{ACM-Reference-Format}
\bibliography{sample-bibliography}

\clearpage

\appendix
\section{LLM data format}

\begin{figure}
  \centering
  \includegraphics[width =\linewidth]{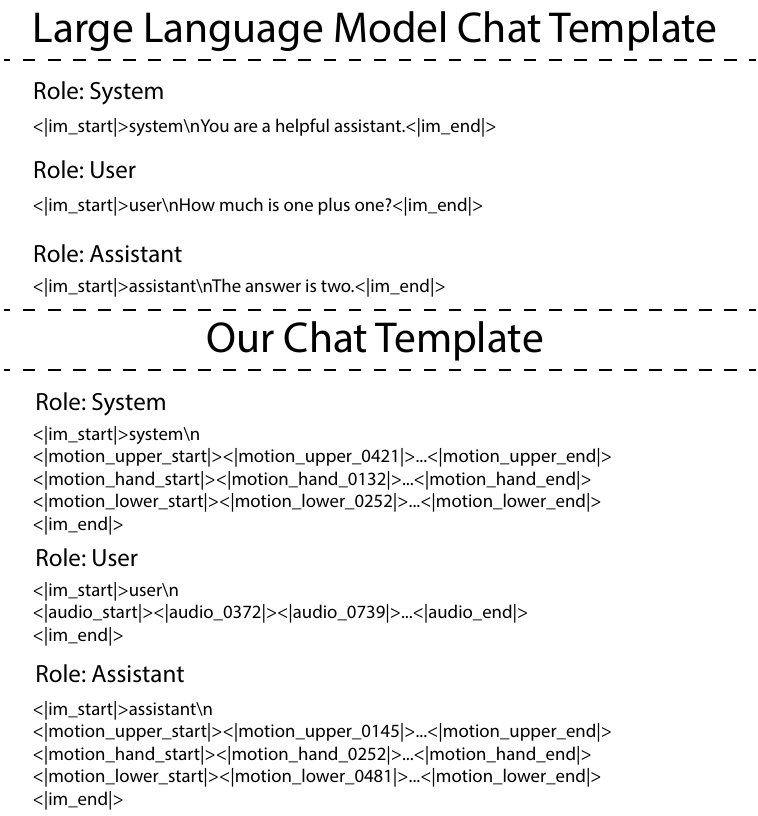}
  \caption{The prompt format of regular LLM and of our method. We prompt the LLM using tokens from motion and audio modals with different template designs.}
  \label{fig:prompt}
\end{figure}

To help readers better understand how our data is organized and fed into the LLM, we visually present the data format in Figure~\ref{fig:prompt}.

\section{Motion Representation}

To validate the effectiveness of introducing residual quantization layers and using only base quantization layer in the subsequent process, we conduct two ablation study in reconstruction task and one in downstream task. As shown in Table~\ref{table1}, recons (w/ res. in train\&infer) refers to the setting where the quantized residual layer is used in both the training and inference stages of the reconstruction task. recons (w/ res. in train) indicates that the quantized residual layer is applied only during training, while recons (w/o res. in train\&infer) denotes that it is not used in either training or inference. The results demonstrate that introducing residual layers in training significantly improves the model's reconstruction performance. Moreover, using only the base quantization layer during inference yields comparable results to using all residual quantization layers. Therefore, in subsequent processes, we only model the base quantization layer, instead of modeling all residual quantization layers~\cite{Zhang2024SemanticGesture,guo2023momask}. It can be also observed that incorporating the quantized residual layer during the training stage of VQ-VAE not only improves reconstruction quality but also enhances the performance on downstream tasks from the comparison between MECo and MECo(w/o res. in train\&infer).

It is worth noting that the setting(w/o res. in train\&infer) results in relatively high BC scores, which, however, do not reflect better quality. From a visual perspective, this is primarily due to the presence of numerous meaningless jitters compared to our default setting(w/ res. in train).

\section{Comparisons with other LLM-based methods}\label{supp:more_compare}

We perform comparisons with relevant techniques that synthesize motions with speech or motion examples as input.

\begin{table}[b]

\centering
\caption{Abalation study on BEAT2~\cite{liu2024emage}. We report FGD $\times 10^{-1}$, BC $\times 10^{-1}$, and diversity.}
{
\begin{tabular}{lccc}
\toprule
     Method & FGD $\downarrow$ & BC $\uparrow$ & Diversity~$\uparrow$   \\ 
\midrule
GT & 0.000 & 6.897 & 12.75 \\

recons~\cbhb{(w/ res. in train\&infer)} & 2.221 & 7.544 & 12.26\\
recons~\cbhb{(w/ res. in train)} & 2.311 & 7.779 & 12.66\\
recons~\cbhb{(w/o res. in train\&infer)} & 2.974 & 7.940 & 12.48\\

\midrule
MECo~ & 3.401 & 7.346  & 15.30 \\
MECo~\cbhb{(w/o res. in train\&infer)} & 3.762 & 7.833  & 15.19 \\
\bottomrule
\end{tabular}}
\label{table1}
\end{table}

\begin{table*}[h]
\caption{Impact of finetuning on LLMs' original text capabilities, M\&S refers to motion and speech.}
\centering
\begin{tabular}{lccccc}
\toprule
Modality & Model Name & Base Model & Original MMLU$\uparrow$ & Finetuned MMLU$\uparrow$ & Degradation$\downarrow$ \\
\midrule
Motion & MotionGPT & flan-t5-base & 33.44 & 22.95 & 31.37\% \\
Speech & SpeechGPT & llama-13b & 46.90 & 27.13 & 42.15\% \\
Vision & QwenVL2.5-7b-instruct & Qwen2.5-7b-instruct & 74.20 & 70.17 & 5.43\% \\
M\&S & MECo & Qwen2.5-0.5b-instruct & 46.50 & 46.27 & 0.49\% \\
3M\&S & MECo(7b llm) & Qwen2.5-7b-instruct & 74.20 & 74.13 & 0.09\% \\
\bottomrule
\end{tabular}

\label{table:mmlu_comparison}
\end{table*}

\subsection{MotionGPT}

MotionGPT~\cite{jiang2024motiongpt} focuses on aligning motion modal with text modal. It does not incorporate speech or motion example control.

As MotionGPT is trained on a diverse range (15) of text-motion related tasks, we designed two experiments: (a) whether MotionGPT supports example-based control though not explicitly trained; (b) adding speech modality to MotionGPT.

When motion example and text description are fed into the MotionGPT model, instead of generating example-guided motion, it returned a textual description. This highlights that enabling example-based control requires specifically designed approaches.

We attempted to introduce speech modality to MotionGPT by constructing a speech-to-gesture task using speech-gesture data pairs from BEAT2. We incorporated this task into finetuning process to enable direct comparison with our model. Since the MotionGPT generated results do not include hands, we assign ground truth values for the hands. The result's FGD is 0.8784, which is much weaker than our 0.3401. Though this may be caused by many factors, this result highlights that different motion generation tasks in different modalities require unique design choices when utilizing LLMs, and it is not easy to transfer a model to another task.

\subsection{T2M-GPT}
T2M-GPT~\cite{zhang2023generating} is a model solely trained on text-to-motion model, which does not support speech or example input. In addition, it does not use pre-trained LLM and was trained from scratch using a GPT structure similar to LLMs.

We designed an experiment to explore whether the T2M-GPT architecture and training design can tackle speech-to-gesture generation. To suit its architecture design, we use the speech transcripts, along with speech tokens as input, and train the model to generate the motion tokens. The final model achieved a FGD score of 0.7253, which is slightly worse than TalkSHOW's results and falls behind our 0.3401. Note that their design does not further allow example-based control.

\subsection{M$^3$GPT}

M$^3$GPT~\cite{luo2024m3gpt} adds music-dance tasks based on MotionGPT, and does not support example-based control. Unfortunately, this work is not open-sourced. Their github repository contains a template without training or inference code, and no model checkpoints are available.

\section{Aligning with text}
We additionally align the text modality during our training process like MotionGPT and M³GPT, adding speech-to-text and text-to-gesture tasks, where the text corresponds to the speech transcripts. However, experimental results indicate that this negatively impacts speech-to-gesture performance, producing the FGD score of 0.4104, while also degrading the original textual capabilities of the LLM, lowering MMLU to 43.73. As we focus on speech-to-gesture solely, aligning with text modal ultimately compromises our primary objective and harms the fundamental capabilities of the LLM.

\section{Impact of finetuning on LLM's original text capabilities}
Regarding whether finetuning compromises LLM's original text capabilities, we provide several examples. Note that for motion related tasks, we only demonstrate MotionGPT, as T2M-GPT does not use pre-trained LLM and M³GPT is not open-sourced.

As shown in Table ~\ref{table:mmlu_comparison}, when incorporating new modalities, existing works generally experience an inevitable degradation in text capabilities, while our method has small impact on the LLM's original text capabilities.

\section{Objective Metrics}
We follow BEAT2~\cite{liu2024emage} benchmark and use the same way to calculate these metric.

\subsection{Fr\'echet Gesture Distance (FGD)}

A lower FGD, as referenced by \cite{yoon2020speech}, indicates that the distribution between the ground truth and generated body gestures is closer. It is currently the metric that most closely aligns with human perception in evaluating the quality of gestures~\cite{genea2022}. Similar to the perceptual loss used in image generation tasks, FGD is calculated based on latent features extracted by a pretrained network:

\begin{equation}
\text{FGD}(g, \hat{g}) = \| \mu_r - \mu_g \|^2 + \text{Tr} \left( \Sigma_r + \Sigma_g - 2 (\Sigma_r \Sigma_g)^{1/2} \right),
\end{equation}

where $\mu_r$ and $\Sigma_r$ represent the first and second moments of the latent features distribution $z_r$ of real human gestures $g$, and $\mu_g$ and $\Sigma_g$ represent the first and second moments of the latent features distribution $z_g$ of generated gestures $\hat{g}$.

\subsection{Beat Constancy (BC)}
A higher BC, suggests a closer alignment between the beat of gesture 
and the speech audio. It can be calculated by:
\begin{equation}
\mathrm{BC} = \frac{1}{g} \sum_{b_g \in g} \exp\left( -\frac{\min_{b_a \in a} \|b_g - b_a\|^2}{2\sigma^2} \right),
\end{equation}

\subsection{L1 Diversity}
A higher diversity indicates a larger variance in the given gesture clips. We calculate the average L1 distance from different N motion clips as follows:
\begin{equation}
\text{L1 div.} = \frac{1}{2N(N-1)} \sum_{t=1}^{N} \sum_{j=1}^{N} \left\| p_t^i - \hat{p}_t^j \right\|_1,
\end{equation}
where $p_t$ represents the position of joints in frame t, note that the character's translation is set to zero.

\subsection{User Study}
\begin{figure}[t]
  \centering
  \includegraphics[width =0.99\linewidth]{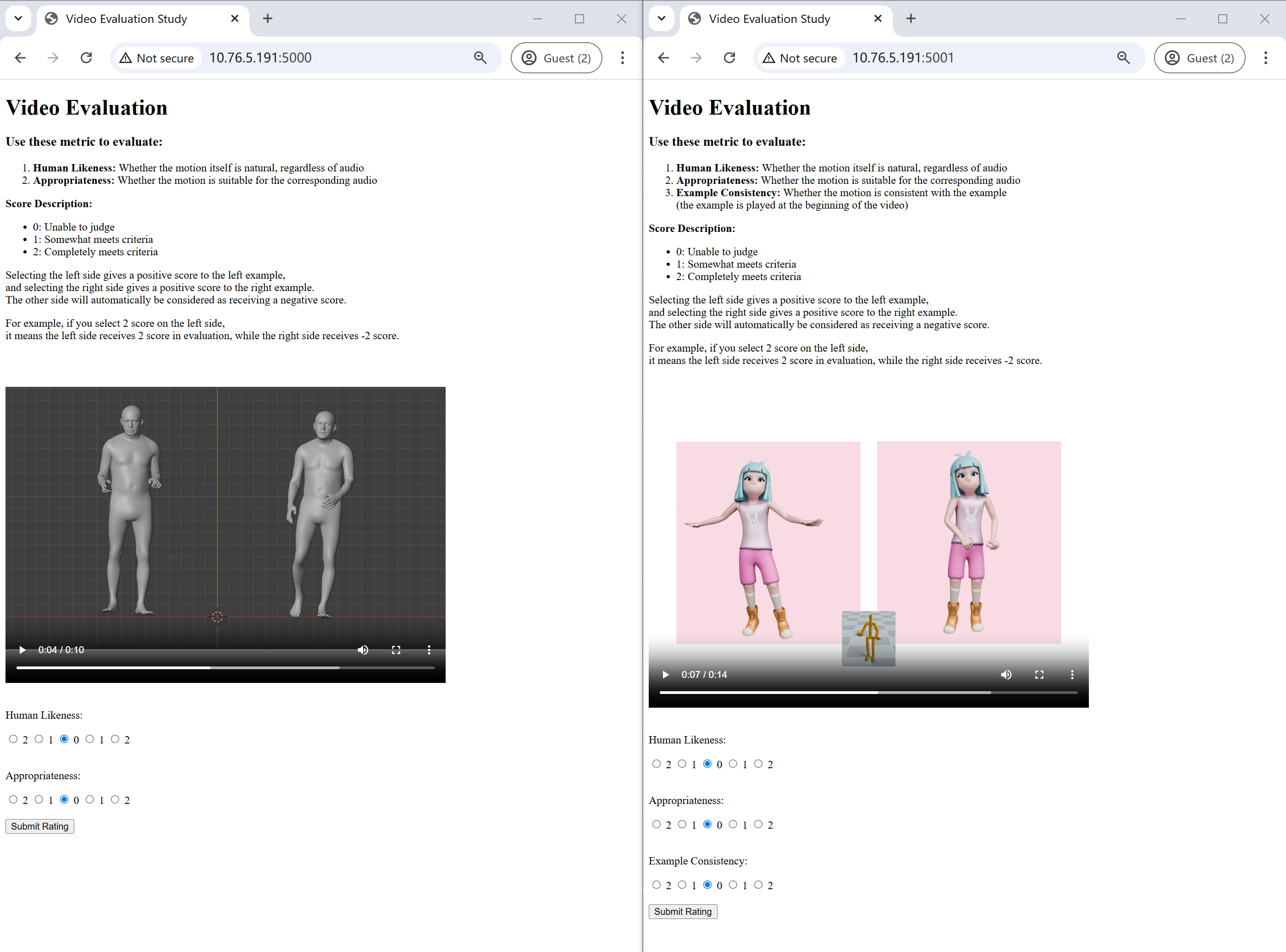}
  \caption{Screenshot of the user interface used for user study.}
  \label{fig:website_shot}
\end{figure}

We conducted the user study on webpages to collect data. The user interface is shown in Figure~\ref{fig:website_shot}. Note that we use two types of avatars during the process: one is the SMPL-X model, and the other is the Amy model from Mixamo. The main reason for using different avatars is that we utilize two datasets during our process: BEAT2 and ZeroEGGS. The BEAT2 dataset is represented in SMPL-X format. Retargeting the hand skeleton from SMPL-X to the Mixamo model is difficult and often produces visually obvious artifacts. Therefore, we directly used the mesh from SMPL-X.

Following GestureDiffuCLIP~\cite{Ao2023GestureDiffuCLIP}, for comparison purposes, we created two webpages to collect results. One evaluates the Human Likeness and Appropriateness of generated motions based solely on audio input. The other evaluates Human Likeness, Appropriateness, and Example Consistency when both audio and motion examples are provided as input. Each webpage is consist of three parts: textual explanation of the evaluation, evaluation videos, and scoring buttons. Each comparison group contains only two cases, which are played simultaneously on the left and right sides of the video for easier comparison.

\section{More Discussion}
Since we first compress the motion into a latent representation using a motion tokenizer, our method struggles to provide joint-level control, such as precisely controlling a character's trajectory. Providing more precise and fine-grained control remains an area worthy of exploration. Additionally, our method can produce up to 36 seconds of motion within 1 second of processing time, but it's important to note that this is offline generation, as our motion tokenizer is not causal—meaning the current motion is influenced by both past and future tokens. A straightforward solution is using a causal motion tokenizer~\cite{jiang2024causal}, which can ensure that current motions are only influenced by past tokens, thereby enabling real-time generation.
\end{document}